\documentclass[10pt]{article}
\usepackage{arxiv}
\usepackage[utf8]{inputenc} % allow utf-8 input
\usepackage[T1]{fontenc}    % use 8-bit T1 fonts
\usepackage{graphicx}
\usepackage[mathlines]{lineno}
\usepackage{amsmath,amssymb,amsthm}
\usepackage{mathtools}
\usepackage{xspace}
\usepackage{epsfig}
\usepackage[lined,noend,linesnumbered,ruled,titlenotnumbered]{algorithm2e}
\usepackage{color}
\usepackage{url}
\usepackage[square,numbers,compress,sort]{natbib}
\usepackage{tikz}
\usepackage{float}
\usepackage{booktabs}
\usepackage{makecell}
\usepackage{multicol}
\usepackage{multirow}
\usepackage{longtable}
\usepackage{array}
\usepackage[libertine,vvarbb]{newtxmath}
\usepackage{bm}
\usepackage{colortbl,soul,xcolor,bbm}
\usepackage{subcaption}
\usepackage{nicefrac}
\usepackage{microtype}      % microtypography
\usepackage[inline]{enumitem}
\usepackage[hang,flushmargin]{footmisc}

\DeclareMathOperator*{\argmin}{\arg\!\min}
\DeclareMathOperator*{\Int}{Int}
\DeclareMathOperator*{\Conv}{Conv}
\DeclareMathOperator*{\poly}{poly}

\setlist[itemize]{align=parleft,left=1pt..1em}
\SetKwInput{Input}{Inputs}
\SetKwComment{tcc}{//}{}

\SetCommentSty{mycommfont}
\newtheorem{lemma}{Lemma}%[chapter]
\newtheorem{corollary}{Corollary}%[chapter]
\newtheorem{definition}{Definition}%[chapter]
\newtheorem{observation}{Observation}%[chapter]

\usepackage[createShortEnv,conf={external appendix}]{proof-at-the-end}
\pratendSetGlobal{text link={}}

\begin{document}
\author{%Authors-affiliations mapping supported
Anh Viet Do$^1$ \and
Aneta Neumann$^1$ \and
Frank Neumann$^1$ \And
Andrew M.\ Sutton$^2$ 
\affiliations
$^1$Optimisation and Logistics, School of Computer and Mathematical Sciences, The University of Adelaide, Adelaide, Australia\\
$^2$Department of Computer Science, University of Minnesota Duluth
}
% \author{%
%   Anh Viet Do\\
%   Optimisation and Logistics\\
%   School of Computer and Mathematical Sciences\\
%   The University of Adelaide\\
%   Adelaide, Australia\\
%   % \texttt{vietanh.do@adelaide.edu.au} \\
%   % examples of more authors
%   \And
%   Andrew M Sutton\\
%   Department of Computer Science\\
%   University of Minnesota Duluth\\
%   % \texttt{email} \\
%   \And
%   Aneta Neumann\\
%   Optimisation and Logistics\\
%   School of Computer and Mathematical Sciences\\
%   The University of Adelaide\\
%   Adelaide, Australia
%   % \texttt{email} \\
%   \And
%   Frank Neumann\\
%  Optimisation and Logistics\\
%   School of Computer and Mathematical Sciences\\
%   The University of Adelaide\\
%   Adelaide, Australia\\
%   % \texttt{email} \\
% }
\title{Rigorous Runtime Analysis of MOEA/D for Solving Multi-Objective Minimum Weight Base Problems}
\renewcommand{\shorttitle}{Runtime Analysis of MOEA/D for Multi-objective Minimum Weight Base Problems}
\date{}
\maketitle
\begin{abstract}
We study the multi-objective minimum weight base problem, an abstraction of classical NP-hard combinatorial problems such as the multi-objective minimum spanning tree problem. We prove some important properties of the convex hull of the non-dominated front, such as its approximation quality and an upper bound on the number of extreme points. Using these properties, we give the first run-time analysis of the MOEA/D algorithm for this problem, an evolutionary algorithm that effectively optimizes by decomposing the objectives into single-objective components. We show that the MOEA/D, given an appropriate decomposition setting, finds all extreme points within expected fixed-parameter polynomial time in the oracle model, the parameter being the number of objectives. Experiments are conducted on random bi-objective minimum spanning tree instances, and the results agree with our theoretical findings. Furthermore, compared with a previously studied evolutionary algorithm for the problem GSEMO, MOEA/D finds all extreme points much faster across all instances.
\end{abstract}
%Keywords:minimum weight base problem, multi-objective optimization, evolutionary algorithm

\section{Introduction}
Evolutionary algorithms have been widely used to tackle
multi-objective optimization problems in many areas such as robotics,
pattern recognition, data mining, bioinformatics, scheduling and
planning, and neural network training \cite{Zhou2011}. Their
population-based search operators make them a natural choice for
simultaneously handling several possibly conflicting objectives. Many generic evolutionary multi-objective frameworks have been developed to supply basic implementations for any problem, and to provide templates that can be fine-tuned for specific applications (we refer to \cite{Wang2023} for an overview of common approaches). Such features, along with their strong empirical performances in challenging applications, have led them to becoming one of the most attractive topics to researchers and practitioners alike.

Among evolutionary multi-objective algorithms (EMOs), arguably the
most exemplary are dominance-based approaches such as GSEMO and NSGA
variants, with the former often being considered a baseline. Another
popular technique for multi-objective optimization is to decompose the
multiple objectives into a single-objective subproblem. The MOEA/D
algorithm is a state-of-the-art application of this technique in
evolutionary computation \cite{Trivedi2016,Xu2020}.

Despite the prevalence of EMOs on practical applications, rigorous
analyses of their runtime behavior on meaningful problems are
scarce. Nevertheless, these kinds of analyses are critical for (1)
providing performance guarantees and guidelines to practitioners who
use and develop these techniques in the field, and (2) promoting the
explainability of heuristic search and optimization techniques by
clarifying their working principles through a careful mathematical analysis.
Run-time analyses on the performance of evolutionary algorithms have been provided for simple algorithms such as GSEMO in both artificial benchmark problems \cite{Bian2018,Doerr2021} and others such as bi-objective minimum spanning tree \cite{Neumann20071,Roostapour2020} and constrained submodular optimization \cite{NIPS2015_b4d168b4,NIPS2017_d7a84628,Do2020,Qian2020}. In recent years, theoretical analyses of state of the art approaches such as NSGA-II and MOEA/D have been conducted \cite{Huang2019,Huang2021,Huang20211,Zheng2022,Doerr2023,cerf2023proven}. Most of these run-time results are on artificial benchmark problems, and the one for NSGA-II on bi-objective minimum spanning tree proves promising.

In this paper, we present for the first time rigorous results on
MOEA/D for a classical multi-objective optimization problem, namely
the multi-objective minimum weight base problem. This problem, falling under the matroid optimization category, is a significant generalization of the previously studied bi-objective minimum spanning tree problem. In this work, we focus on approximating the non-dominated front, as determining whether the front is reached is EXPSPACE. In particular, we show that MOEA/D obtains a factor $2$-approximation for two objectives in expected polynomial time. Previous analyses for the special case of graphic matroid (i.e. spanning forests) were only able to show a pseudo-polynomial run-time for GSEMO to obtain this approximation \cite{Neumann20071}. We further extend the analyses by deriving a fixed-parameter polynomial expected run-time in instances with $k>2$ objectives to reach a $k$-approximation.

Instrumental to our analyses is a deeper understanding of the problem, and as such, we formally examine certain properties of the multi-objective minimum weight base problem. We first prove a tight approximation guarantee from computing the convex hull of the non-dominated front, extending the known guarantee for two objectives \cite{Neumann20071}. With this in mind, we explore insight regarding this convex hull, including its vertex complexity and the structural relation among solutions whose weights constitute said convex hull. In addition, we briefly formulate an efficient deterministic approach to enumerate extreme points. These findings may be of interest in areas beyond runtime analysis.

\section{Preliminaries \& Problem}
First, we give an overview of relevant matroid theory concepts, with terminologies adopted from the
well-known text book \cite{Oxley2011} on the subject.

\begin{definition}\label{def:matroid}
A tuple $M=(E,\mathcal{I}\subseteq2^E)$ is a \emph{matroid} if \begin{enumerate*}[label=\itshape\alph*\upshape)]
\item $\emptyset\in\mathcal{I}$,
\item $\forall x\subseteq y\subseteq E, y\in\mathcal{I}\implies x\in\mathcal{I}$,
\item $\forall x,y\in\mathcal{I},\allowbreak|x|<|y|\implies\exists e\in y\setminus x,x\cup\{e\}\in\mathcal{I}$.
\end{enumerate*} The set $E$ is the ground set, and $\mathcal{I}$ is the independence collection. A \emph{base} of $M$ is a maximal set in $\mathcal{I}$.
\end{definition}
\begin{definition}
Given a matroid $M=(E,\mathcal{I})$, its \emph{rank function}, $r:2^V\to\mathbb{N}$, is defined as $r(x)=\max\{|y|:y\in 2^x\cap\mathcal{I}\}$, and the rank of $M$ is $r(E)$. A matroid is completely characterized by its rank function.
\end{definition}
To give examples, a $K$-rank uniform matroid over $E$ admits the
independence collection $\mathcal{I}=\{x\subseteq E:|x|\leq K\}$,
characterizing a cardinality constraint. In linear algebra, a
representable matroid describes linear independence among a vector
set. In graph theory, given an undirected graph $G=(V,E)$, a graphic
matroid $M=(E,\mathcal{I})$ defined by $G$ is such that $\mathcal{I}$
contains all edge sets $x$ forming a forest subgraph in
$G$. A base of a graphic matroid is a spanning forest, which itself is an object of much interest. Dual to the graphic matroid, the bond matroid $M^*=(E,\mathcal{I}^*)$ is such that $\mathcal{I}^*$ contains all edge sets $x$ whose removal from $E$ preserves every pairwise connectivity in $G$. The matroid properties emerge in many combinatorial structures of various optimization problems \cite{Oxley2011}.

A classical application of matroids in optimization is in the minimum weight base (MWB) problem. Given a weighted matroid $(E,r,w)$, this problem asks to find a base in this matroid minimizing $w$. The most arguably well-known special case of MWB problem is the minimum spanning tree (MST) problem. It is known that the classical Greedy algorithm minimizes (and maximizes) arbitrary weight over a base collection of any matroid \cite{Rado1957,Gale1968,Edmonds1971}. From the exchange property between independent sets, we see that Greedy can also enumerate all minimum weight bases, thus characterizes the optimality of any MWB instance.

The multi-objective minimum weight base (MOMWB) is a natural multi-objective extension to MWB. Given a $k$-weighted\footnote{The integrality does not affect the algorithms' behaviors. This assumption is used to ease the analysis. The positivity assumption ensures that approximation factors are meaningful.} matroid
$(E,r,w\in(\mathbb{N}^*)^{k\times|E|})$ where $E$ is the ground set,
$r$ is the rank function of the matroid, and the weight vector of a
solution $x\in\{0,1\}^{|E|}$ is $wx$ (also called the image of $x$ under $w$), the multi-objective problem asks
to find a non-dominated set of bases in $(E,r)$ minimizing $w$. Let
$m:=|E|$, $n:=r(E)$, for $x,y\in\{0,1\}^m$ and a function
$f:\{0,1\}^m\to\mathbb{R}^k$, $x$ dominates $y$, denoted with
$x\preceq_f y$, iff $f(y)-f(x)\in\mathbb{R}^k_{\geq0}$. We see that
$x\preceq_f y$ iff
$\min_{\lambda\in[0,1]^k}\lambda^\intercal(f(y)-f(x))\geq0$, i.e. $y$
has greater scalarized objective value than $x$ across all linear
trade-offs. We denote the set of images of non-dominated solutions with $F$, and vertices of its convex hull $\Conv(F)$ are called \emph{extreme points}. For convenience, let $\Conv(F)$ contain only points in $F$ and that its faces be conventionally defined, i.e. as continuous Euclidean subspaces.

Since $F$ can be exponentially large, we consider instead
approximating it by finding solutions mapped to $\Conv(F)$. Such a set
is known to guarantee a $2$-approximation of the non-dominated set for $k=2$ \cite{Neumann20071} under the following definition.

\begin{definition}[Minimization]
Given $k$ non-negative objective functions $f:=(f_i)_{i=1}^k$, a solution $x$ \emph{$c$-approximates} a solution $y$ for some $c\geq0$ if $cf(y)-f(x)\in\mathbb{R}^k_{\geq0}$. A solution set $X$ \emph{$c$-approximates} (or is a $c$-approximation of) a solution set $Y$ if every $y\in Y$ is \emph{$c$-approximated} by at least a $x\in X$.
\end{definition}

We formally describe categories of solutions of interest. Here, we only consider feasible solutions, e.g. bases in a MWB or MOMWB instance. Furthermore, a subset of $E$ is characterized by a bit-string in $\{0,1\}^{|E|}$, so both set and bit operations on solutions are well-defined, and we use both representations throughout the paper.
\begin{definition}
A solution $x$ is a \emph{supported solution} to an instance with objective functions $f=(f_i)_{i=1}^k$ and a solution set $S$ if there is a linear trade-off $\lambda\in[0,1]^k\setminus\{\bm{0}\}$ where $x\in\argmin_{y\in S}\lambda^\intercal f(y)$. A trade-off set $\Lambda$ is \emph{complete} if $\bigcup_{\lambda\in\Lambda}\argmin_{y\in S}\lambda^\intercal f(y)$ contains all supported solutions. A supported solution $z$ is \emph{extreme} if there is $\lambda'\in[0,1]^k$ where for all $x\in\argmin_{y\in S}\lambda'^\intercal f(y)$, $f(z)=f(x)$. A set containing a trade-off for each extreme solution is called \emph{sufficient}.
\end{definition}

We see that supported solutions are precisely the solutions whose images lie on $\Conv(F)$. Intuitively, a complete trade-off set decomposes the multi-objective instance in such a way to allow enumerating all supported solutions via exactly solving scalarized instances. Since supported solutions that are not extreme are mapped to points on the faces of $\Conv(F)$, we have the following observation.
\begin{observation}\label{observation:extreme_sufficient}
For each supported solution minimizing $\lambda^\intercal f$, there is an extreme solution minimizing $\lambda^\intercal f$. For every $\lambda\in[0,1]^k$, there is an extreme solution minimizing $\lambda^\intercal f$.
\end{observation}

However, for linear functions, the number of supported solutions can
be very large, so we also consider finding a representative subset which, as we will show, is sufficient to give an approximation guarantee.
\begin{definition}
A solution set $X$ is sufficient to an instance with
objective functions $f$ if for every extreme solution $y$, there is
$x\in X$ where $f(x)=f(y)$. The analogy for supported solutions is
called a complete solution set.
\end{definition}

With this definition, the set of solutions that are mapped to the extreme points is sufficient. In fact, the size of a minimal sufficient set is exactly the number of extreme points. Note that while the set of all supported solutions is unique, there can be multiple distinct minimal sufficient sets due to duplicate images. We briefly prove an approximation factor by any sufficient set, which is not restricted to MOMWB.

\begin{theoremE}[Minimization][category=convf,end]\label{theorem:k-approx}
Given $k\geq1$ and a $k$-objective instance admitting a non-negative feasible minimum in each objective and a sufficient solution set $P$, $P$ $k$-approximates all solutions. This factor is tight for all $k$, even if $P$ is a complete solution set.
\end{theoremE}
\begin{proof}
Let $f:=(f_i)_{i=1}^k$ be the objective function vector, $z$ be any solution, $Z=\{i:f_i(z)=0\}$, if $|Z|=k$ then $z$ is an extreme solution, so $P$ $1$-approximates $z$. Assume otherwise, let $\delta_i$ be the minimum positive $f_i$ value of solutions, we define $\lambda\in(0,1]^k$ where $\lambda_i:=2\epsilon/\delta_i$ if $i\in Z$ and $\lambda_i:=\epsilon/[(k-|Z|)f_i(z)]$ otherwise for some sufficiently small $\epsilon>0$. By definition of sufficient solution set and Observation \ref{observation:extreme_sufficient}, there is $x\in P$ minimizing $\lambda^\intercal f$, i.e. $\lambda^\intercal f(x)\leq\lambda^\intercal f(z)=\epsilon$. If $f_i(x)>0$ for some $i\in Z$ or $f_i(x)>(k-|Z|)f_i(z)$ for some $i\notin Z$, then since $f(x)\in\mathbb{R}^k_{\geq0}$, we have $\lambda^\intercal f(x)>\epsilon$, a contradiction. Therefore, $x$, and by extension $P$, $(k-|Z|)$-approximates $z$. Since $z$ can assume positive values in all objectives\footnote{If $|Z|\geq k'$ for all solutions $z$, the instance is reducible to $(k-k')$-objective instances, and the guarantee factor is likewise tight.}, this factor simplifies to $k$.

We show tightness by construction. Let $\epsilon\in(0,k)$, $m:=k^2$, $\theta_i:=\sum_{j=0}^{k-1}e_{ik-j}$ for $i=1,\ldots,k$ where $e_j$ is the $j$th unit vector in $\mathbb{R}^m$, we define a non-negative $k$-objective instance over $\{0,1\}^m$: $\min_x\{f(x):=(\theta_i^\intercal x-\epsilon\prod_{j=0}^{k-1}x_{ik-j})_{i=1}^k:|x|\geq k\}$. We see that the set of all supported solutions is precisely $S:=\{\theta_i\}_{i=1}^k$. Let $z:=\sum_{i=0}^{k-1}e_{ik+1}$ be a solution, for all $i=1,\ldots,k$, $f_i(\theta_i)=k-\epsilon\geq(k-\epsilon)f_i(z)$ (equality holds if $k>1$). This means $S$ fails to $(k-\epsilon-\varepsilon)$-approximate $z$ for any $\varepsilon>0$, and $\epsilon$ can be arbitrarily small. Since $S$ is a complete solution set, the claim follows.
\end{proof}

We denote the weight scalarization with trade-off $\lambda$ with $w^{(\lambda)}:=\lambda^\intercal w$, so $(E,r,w^{(\lambda)})$ is a scalarized instance at $\lambda$.% All proofs, including the one for the above result, are included in the Appendix.

\section{Properties of Conv($F$) in multi-objective minimum weight base problem}\label{sec:convf}
Here, we derive various properties of $\Conv(F)$ with implications on the problem's approximation complexity. Since these solutions are optima of scalarized instances, we use the properties of the Greedy algorithm, known to guarantee and characterize optimality in linear optimization over a matroid.

The Greedy algorithm starts from an empty set and adds elements to it
in increasing weight order while maintaining its independence, until a
base is reached. In essence, Greedy operates on a permutation over $E$
and produces a unique solution so we can characterize its outputs via
permutations. We say a permutation $\tau$ over $E$ \emph{sorts} the
weight $w$ if, for all $i=1,\ldots,m-1$, $w_{\tau(i)}\leq
w_{\tau(i+1)}$. As mentioned, Greedy run on a permutation that
sorts the weight to be minimized returns a minimum weight base. More importantly, all minimum weight bases can be obtained by running Greedy on all sorting permutations. This allows us to derive properties of any solution mapped onto $\Conv(F)$ using Greedy's behaviors. In particular, we can circumvent the difficulty of examining weights by examining permutations instead, essentially looking at the weights' rankings rather than the weights themselves.

To simplify analysis, we restrict the trade-off space to non-negative $L_1$-normalized vectors $U=\left\lbrace a\in[0,1]^k:\sum_{i=1}^ka_i=1\right\rbrace$, let $\pi_\lambda$ be a permutation sorting $w^{(\lambda)}$ for $\lambda\in U$. For each $i\in E$, let $\mathbb{w}_i=(w_{j,i})_{j=1}^k$, and for each pair $i,j\in E$, let $\delta_{i,j}=\mathbb{w}_i-\mathbb{w}_j$ and $\Delta_{i,j}=\left\lbrace a\in U:\delta_{i,j}^\intercal a=0\right\rbrace$ be the $(k-2)$-dimensional set characterized by the fact that for all $\lambda\in U$, $w^{(\lambda)}_i=w^{(\lambda)}_j$ iff $\lambda\in\Delta_{i,j}$. Finally, let $A$ be the multiset of non-empty $\Delta_{i,j}$ where $\delta_{i,j}\neq0$, $H_A$ be the multiset of convex $(k-1)$-polytopes in $U$ defined by intersections of half-spaces bounded by hyperplanes in $A$ and the boundary of $U$, and $A'$ be the set of points in $U$ where each point lies in the interior of a polytope in $H_A$, we show that $A$ and $A'$ encompass complete solution set and sufficient solution set, respectively. Note that if $\delta_{i,j}=\mathbb{w}_i-\mathbb{w}_j=0$, the inclusion of either $i$ or $j$ in a solution does not change its image under $w$.\footnote{We include polytopes with empty interiors in $H_A$ to account for overlapping hyperplanes in $A$. Furthermore, we assume these hyperplanes are ordered arbitrarily along their normal direction for the purpose of defining interior-free polytopes: $h$ such hyperplanes form $h-1$ polytopes.}

\begin{lemmaE}[][category=convf,end]\label{lemma:tie_partition}
For any $Q\in H_A$, the set of all bases minimizing $w^{(\lambda)}$
remains constant for all $\lambda\in\Int(Q)$\footnote{Given a set $A$ in a metric space, $\Int(A)$ is the set of its interior points.}, and these bases
share an image under $w$. Furthermore, they also minimize
$w^{(\lambda)}$ for all $\lambda\in Q$.
\end{lemmaE}
\begin{proof}
Let $c$ be any point in $\Int(Q)$, by definition of $A$, $w^{(c)}_i=w^{(c)}_j$ iff $\delta_{i,j}=0$, and $w^{(c)}$ admits multiple minima iff they contain different elements among those sharing weights in $w^{(c)}$, while sharing all other elements. Indeed, let $x$ and $y$ be a pair of minima violating this condition, they must contain different sets of weights so for all bijection $\gamma$ between $x\setminus y$ and $y\setminus x$, there is $u\in x\setminus y$ where $w^{(c)}_u\neq w^{(c)}_{\gamma(u)}$; this leads to a contradiction when combined with the base exchange property. This means these optima share image under $w$, and bases not having the same image do not minimize $w^{(c)}$.

Let $b$ be any point on the boundary of $Q$ and $L$ be the set of points between $b$ and $c$ excluding endpoints, we show that $\pi_c$ also sorts $w^{(b)}$. Let $i.j\in E$ where $w^{(c)}_i<w^{(c)}_j$, then $w^{(b)}_i>w^{(b)}_j$ implies $w^{(d)}_i=w^{(d)}_j$ for some $d\in L$, meaning $L$ meets a hyperplane in $A$, a contradiction as $L\subseteq\Int(Q)$. For all pairs $i,j\in E$ where $w^{(c)}_i=w^{(c)}_j$, $\delta_{i,j}=0$ so $w^{(b)}_i=w^{(b)}_j$. With this, every pair is accounted for, so $\pi_c$ sorts $w^{(b)}$. Therefore, since Greedy guarantees optimality, any base minimizing $w^{(c)}$ also minimizes $w^{(b)}$, yielding the claim.
\end{proof}
This immediately gives the upper bound on the number of extreme points, which is the maximum the number of space partitions by hyperplanes; the formula for this is given in \cite{Zaslavsky1975}.
\begin{corollaryE}[][category=convf,end]\label{collorary:sufficient_no_upperbound}
The size of a minimal sufficient solution set is at most $\sum_{i=1}^{k}{m(m-1)/2\choose i-1}$, and $A'$ is a sufficient trade-off set.
\end{corollaryE}
\begin{proof}
We see that $|A'|=|H_A|$, which is upper bounded by the number of half-space intersections from hyperplanes in $A$. Since these are $(k-2)$-dimensional hyperplanes, applying the formula in \cite{Zaslavsky1975} gives $|H_A|\leq\sum_{i=1}^{k}{|A|\choose i-1}$ which is increasing in $|A|$, so the claim follows from $|A|\leq m(m-1)/2$. We have $A'$ is a sufficient trade-off set following from Lemma \ref{lemma:tie_partition} and $\bigcup_{Q\in H_A}Q=U$.
\end{proof}

We remark that we deliberately choose each trade-off in $A'$ from the interior of each polytope. This is because if a zero trade-off coefficient is assigned to an objective, then bases minimizing the weight scalarized by such a trade-off may not be non-dominated. Furthermore, such scalarized weights admit optima whose images under $w$ are identical, which is necessary to ensure that the first optimum an optimization algorithm finds using these trade-offs is an extreme solution. Moreover, this trade-off selection scheme also guarantees that said algorithm does not discard vertices of $\Conv(F)$ over time, unless it stores all found optima.

Given a solution set $S$, the $l$-Hamming neighborhood graph of $S$ is an undirected graph $G_l=(S,\{\{a,b\}:|a\otimes b|\leq l\})$, and $S$ is $l$-Hamming connected if $G_l$ is connected. Neumann \cite{Neumann20071} proved for spanning trees that given the
non-dominated front being strongly convex, the set of supported solutions is $2$-Hamming connected. We show that this even holds for matroid bases without the convexity assumption. For simplicity, we assume, for the rest of the analysis, fixed orderings among each class of elements $i\in E$ sharing $\mathbb{w}_i$. We will see that the existence of such elements does not affect the $2$-Hamming connectivity among supported solutions.

We first show that as the trade-off moves continuously within $U$, the permutation sorting the scalarized weight is transformed incrementally by exchanging two adjacent positions, which we call an adjacent swap.
\begin{lemmaE}[][category=convf,end]\label{lemma:adj_swap_separation}
For any $a,a'\in U$, let $A^*=\{a_i\}_{i=1}^h$ be the multiset of intersections between the line segment connecting $a$ and $a'$ and hyperplanes in $A$, indexed in the order from $a$ to $a'$, there is a shortest sequence of adjacent swaps from $\pi_{a}$ to $\pi_{a'}$, $(\pi_a,\tau_1,\ldots,\tau_h,\pi_{a'})$, where for all $i=1,\ldots,h$, $\tau_i$ sorts $w^{(a_i)}$. If $w^{(a)}$ or $w^{(a')}$ can be sorted by multiple permutations, the claim holds assuming that $\pi_{a}$ and $\pi_{a'}$ have maximum Kendall distance\footnote{Kendall distance between two permutations equals the minimum number of adjacent swaps needed to transform one into the other \cite{Kendall1938}.}.
\end{lemmaE}
\begin{proof}
Let $0<\lambda_c,<\lambda_d<1$ such that $b:=(1-b)a+ba'\in A^*$ and for all $\lambda\in[\lambda_c,\lambda_b)$, $(1-\lambda)a+\lambda a'\notin A^*$, and let $c:=(1-\lambda_c)a+\lambda_c a'$, then elements sharing weight in $w^{(b)}$ must be mapped to consecutive positions in $\pi_c$. Indeed, let $p,q\in E$ ($\pi_{c}(p)<\pi_{c}(q)$) where $w^{(b)}_{p}=w^{(b)}_{q}$, if there is $o\in E$ where $\pi_c(o)\in(\pi_c(p),\pi_c(q))$ and $w^{(b)}_{o}\neq w^{(b)}_{p}$, then since the former implies $w^{(c)}_{o}\in(w^{(c)}_{p},w^{(c)}_{q})$, we have $w^{(d)}_{o}=w^{(d)}_{p}$ or $w^{(d)}_{o}=w^{(d)}_{q}$ for some $d$ in the open line segment connecting $b$ and $c$ which implies $d\in A^*$, a contradiction. Each such consecutive sequence of $l$ positions contains $l(l-1)/2$ pairs. From here, we consider two cases:
\begin{itemize}
\item If such a sequence contains no pair $(i,j)$ where $\delta_{i,j}=0$, then the aforementioned pairs correspond to $l(l-1)/2$ duplicates of $b$ in $A^*$. Furthermore, since the weights are transformed linearly w.r.t. trade-off, for all sufficiently small $\epsilon>0$, these sequences are reversed between $\pi_c$ and $\pi_{b+\epsilon(b-c)}$, whereas positions not in these sequences are stationary. Reversing $l$ consecutive positions requires $l(l-1)/2$ adjacent swaps, so the Kendall distance between $\pi_c$ and $\pi_{b+\epsilon(b-c)}$ equals the multiplicity of $b$ in $A^*$.
\item If such a sequence contains $h>1$ elements with the same weight at all trade-off, then these must occupy consecutive positions in $\pi_c$. As we assumed, the relative ordering among these elements is fixed, so exactly $h(h-1)/2$ swaps are saved. Furthermore, any pair $(i,j)$ among these elements is such that $\Delta_{i,j}\notin A$, meaning these $h(h-1)/2$ pairs are already subtracted from $A^*$.
\end{itemize}
In any case, we can assign to each duplicate of $b$ in $A^*$ a permutation sorting $w^{(b)}$ so that these form a sequence of adjacent swap from $\pi_c$ to $\pi_{b+\epsilon(b-c)}$ including $\pi_{b+\epsilon(b-c)}$ and not $\pi_c$. This directly yields the claim if $a$ and $a^*$ are not in $A^*$.

Assume otherwise, then for all hyperplanes $\Delta_{i,j}$ containing $a$, $w^{(a)}_i=w^{(a)}_j$, so for every such pair $(i,j)$, we arrange $\pi_a$ so that their pairwise ordering in $\pi_a$ is the reverse of that in $\pi_{a'}$. We likewise give $a'$ the same treatment\footnote{This is also done for any $\Delta_{i,j}$ containing both $a$ and $a'$.}. With this, the Kendall distance between $\pi_{a}$ and $\pi_{a'}$ is maximized and equal to $|A^*|$.
\end{proof}

Next, we show that an adjacent swap on the sorting permutation incurs an at most 2-bit change in the minimum weight base.
\begin{lemmaE}[][category=convf,end]\label{lemma:adj_swap_hamming}
Let $\tau$ and $\tau'$ be permutations over $E$ that are one adjacent swap apart, and $x$ and $x'$ are Greedy solutions on them, respectively, then $|x\otimes x'|\leq2$. Furthermore, let $u,v\in E$ where $\tau(v)=\tau(u)+1$ and $\tau'(v)=\tau'(u)-1$, $|x\otimes x'|=2$ iff $\{u\}=x\setminus x'$ and $\{v\}=x'\setminus x$.
\end{lemmaE}
\begin{proof}
Let $E_o:=\{a\in E:\tau(a)<\tau(o)\}$ be the set of elements Greedy considers adding to $x$ before $o\in E$ when run on $\tau$, we have $x\cap E_u=x'\cap E_u$. If $v\in x$ or $v\notin x'$ or $u\notin x$ or $u\in x'$ then $x=x'$, as can be seen from how Greedy selects elements:
\begin{itemize}
\item If $v\in x$, then $v\in x'$ since Greedy considers $v$ before $u$ when run on $\tau'$. Whether Greedy adds $u$ to $x$ only depends on whether there is a circuit in $(x\cap E_u)\cup\{u\}=(x'\cap E_u)\cup\{u\}$, so it makes the same decision when run on $\tau'$. Afterwards, it proceeds identically on both permutations, leading to the same outcome, so $x=x'$. By symmetry, the same follows from $u\in x'$.
\item If $u\notin x$, then there is a circuit in $(x\cap E_u)\cup\{u\}=(x'\cap E_u)\cup\{u\}$, so $u\notin x'$. By the same argument, Greedy makes the same decision regarding $v$ on both permutations, leading to $x=x'$. By symmetry, the same follows from $v\notin x'$.
\end{itemize}
Assume otherwise, it is a known property of bases \cite{Oxley2011} that $x\cup\{v\}$ contains a unique circuit $C$ and that $v\in C$. Greedy not adding $v$ to $x$ implies that $C\subseteq(x\cap E_v)\cup\{v\}=(x'\cap E_v)\cup\{u,v\}$. Let $v'$ be the first element after $v$ that $x$ and $x'$ differ at and assume w.l.o.g. $v'\in x\setminus x'$, we have $(x'\cap E_{v'})\cup\{u\}=(x\cap E_{v'})\cup\{v\}$ and since $v'$ is not added into $x'$ before Greedy terminates, there must be another circuit in $(x'\cap E_{v'})\cup\{v'\}\subset x\cup\{v\}$ containing $v'$, which is distinct from the unique circuit $C$. The contradiction implies that $x$ and $x'$ do not differ after $v$, so $x\otimes x'=\{u,v\}$.
\end{proof}

Lemma \ref{lemma:adj_swap_separation} and \ref{lemma:adj_swap_hamming} indicate that there is a sequence of 2-bit flips between any pair of supported solutions such that every step also gives a supported solution. Therefore, starting from a supported solution, we can compute the rest of $\Conv(F)$ with 2-bit variations. Note that for a supported solution $x$ minimizing $w^{(\lambda)}$, if there is a class of equal-weight elements $Z$ partially intersecting $x$, then all supported solutions minimizing $w^{(\lambda)}$ containing different elements in $Z$ can be reached from $x$ by a sequence of 2-bit flips, whose each step produces a supported solution also minimizing $w^{(\lambda)}$. This is because $Z$ is located consecutively in $\pi_\lambda$ and can be arranged arbitrarily (leading to the Greedy solution minimizing $w^{(\lambda)}$), and there is a sequence of adjacent swaps between any two permutations.
\begin{corollary}\label{corollary:2bit_enum}
Given solutions $x$ and $y$ where $wx,wy\in\Conv(F)$, there is a non-empty set of solutions $\{z_i\}_{i=1}^h$ where $x=z_1$, $y=z_h$, $|z_{i}\otimes z_{i+1}|=2$ for all $i=1,\ldots,h-1$ and $\{wz_i\}_{i=1}^h\subseteq\Conv(F)$.
\end{corollary}

Lemma \ref{lemma:adj_swap_hamming} also lets us derive a stronger bound on the number of distinct Greedy solutions as the trade-off moves in a straight line, giving an upper bound on the number of extreme points in case $k=2$.
\begin{theoremE}[][category=convf,end]\label{theorem:2obj_extreme_bound}
Given $n\geq1$, $a,b\in U$ and $X$ is a minimal set of extreme solutions such that for each $\theta\in[0,1]$, $X$ contains a solution minimizing $w^{((1-\theta)a+\theta b)}$, $|X|\leq hm-h(h+1)/2+1$ where $h:=\left\lceil\sqrt{2\min\{n,m-n\}-1}\right\rceil$.
\end{theoremE}
\begin{proof}
We define $l_c:=(1-c)a+cb$ for $c\in[0,1]$, let $0\leq\theta\leq\theta'\leq1$ where $\pi_{l_\theta}$ and $\pi_{l_{\theta'}}$ are an adjacent swap apart\footnote{If $\theta=\theta'$, we assume $\pi_{l_\theta}$ is closer to $\pi_a$ in Kendall distance.} and the Greedy solutions on them, $x$ and $x'$, are such that $|x\otimes x'|=2$. Let $u,v\in E$ where $x\cap\{u,v\}=\{u\}$ and $x'\cap\{u,v\}=\{v\}$, Lemma \ref{lemma:adj_swap_hamming} implies $\pi_{l_\theta}(u)<\pi_{l_\theta}(v)$ and $\pi_{l_{\theta'}}(u)>\pi_{l_{\theta'}}(v)$, so $\pi_a(u)<\pi_a(v)$. This means as the trade-off moves from $a$ to $b$, the Greedy solution minimizing the scalarized weight changes incrementally by having exactly one element shifted to the right on $\pi_a$ (to a position not occupied by the current solution). Since at most $hm-h(h+1)/2$ such changes can be done sequentially, Greedy produces at most $hm-h(h+1)/2+1$ distinct solutions in total across all trade-offs between $a$ and $b$.

To show this upper bound, we keep track of the following variables as the trade-off moves from $a$ to $b$. Since each solution contains $n$ elements, let $p_i$ be the $i$th leftmost position on $\pi_a$ among those occupied by the current Greedy solution for $i=1,\ldots,n$, we see that upon each change, there is at least a $j\in\{1,\ldots,n\}$ where $p_j$ increases. Furthermore, for all $i$, $p_i$ can increase by at most $m-n$ since it cannot be outside of $[i,m-n+i]$, so the quantity $p:=\sum_{i=1}^np_i$ can increase by at most $n(m-n)$. We see that $p$ increases by $l$ when the change is incurred by a swap in the Greedy solution such that the added element is positioned $l$ to the right of the removed element on $\pi_a$, we call this a $l$-move. Furthermore, each element pair participates in at most one swap, so $p$ can be increased by at most $m-l$ $l$-moves for every $l=1\ldots,m-1$. Therefore, to upper bound the number of moves, we can assume smallest possible distance in each move, and the increase in $p$ from using all possible $l$-moves for all $l=1,\ldots,h$ is $\sum_{j=1}^hj(m-j)\geq n(m-n)$. This means no more than $\sum_{j=1}^h(m-j)=hm-h(h+1)/2$ moves can be used to increase $p$ by at most $n(m-n)$.
\end{proof}

\begin{corollary}\label{corollary:2obj_extreme_bound}
A bi-objective MWB instance (i.e. $k=2$) admits at most $O(m\sqrt{\min\{n,m-n\}})$ extreme points.
\end{corollary}

We remark that aside from the trivial case $n=1$, we did not find an instance where this bound is tight. As far as we are aware, it is an open question whether this bound is optimal.

\section{Exact computation of extreme points}
In this section, we describe a deterministic framework that finds a solution for each extreme point, as well as a complete trade-off set. This framework, modified from the algorithm proposed in \cite{Hamacher1994} for bi-objective MST, is outlined in Algorithm \ref{alg:convex_fillkd}. It calls another algorithm (e.g. Greedy) to find MWB to scalarized weights, and iteratively computes new extreme points based on information from previous ones. Intuitively, each subset of extreme points $Z$ is such that its convex hull is ``inside'' $\Conv(F)$ and contains, for each extreme point $y\notin Z$, a facet separating $Z$ from $y$. This means $y$ can be discovered by minimizing the weight scalarized along the normal direction of this facet, essentially expanding $\Conv(Z)$ to ``touch'' $y$. This iterative process begins with an optimum in each objective, and ends when all new normal vectors are duplicates of ones found previously, indicating that the current convex hull cannot be expanded further and equals $\Conv(F)$. The special case of this algorithm for $k=2$ is given in Algorithm \ref{alg:convex_fill2d} which treats trade-offs as scalars.

Algorithm \ref{alg:convex_fillkd} requires $O(\#(\poly(k)+k!m\log m))$ operations and $O(k!\#m)$ calls to the matroid rank oracle where $\#$ is the number of extreme points. Each iteration in the main loop adds at least one extreme point, and redundant points are excluded from future iterations via $\Lambda'$. Here, we assume updating trade-off for each new vertex takes $\poly(k)$ operations. Note that exhaustive tie-breaking over all objectives is done at line \ref{line:tiebreak} to ensure that the computed points are the vertices of $\Conv(F)$ instead of interior points of its faces, and that all extreme points are accounted for when the termination criterion is met. Furthermore, if the trade-off assigns zero value to some objectives, this also guarantees that the resulted solutions are non-dominated.

\begin{algorithm}[t]
\KwIn{Multi-weighted matroid $(E,r,w)$}
\KwOut{$S$, $\Lambda$}
$S,\Lambda'\gets\emptyset$\;
$\Lambda\gets\{e_i\}_{i=1}^k$\;
$P\gets$ all permutations over $\{1,\ldots,k\}$\;
\While{$\Lambda\setminus\Lambda'\neq\emptyset$}{
\For{$\lambda\in\Lambda\setminus\Lambda'$}{
$\forall p\in P,a_p\gets$ base minimizing $w^{(\lambda)}$ prioritizing weights ranked by $p$\;\label{line:tiebreak}
$S\gets S\cup\{a_p\}_{p\in P}$\;
}
$\Lambda'\gets\Lambda'\cup\Lambda$\;
$\Lambda\gets$ non-negative normal vectors to facets of $\Conv(\{wx:x\in S\})$\;
}
\caption{Finding extreme points and a complete trade-off set (adapted from \cite{Hamacher1994})}
\label{alg:convex_fillkd}
\end{algorithm}

\begin{algorithm}[t]
\KwIn{Bi-weighted matroid $(E,r,w_1,w_2)$}
\KwOut{$S$, $\Lambda$}
$S,\Lambda'\gets\emptyset$\;
$\Lambda\gets\{0,1\}$\;
\While{$\Lambda\setminus\Lambda'\neq\emptyset$}{
\For{$\lambda\in\Lambda\setminus\Lambda'$}{
$a,b\gets$ bases minimizing $(1-\lambda)w_1+\lambda w_2$ prioritizing $w_1$ and $w_2$, respectively\;
$S\gets S\cup\{a,b\}$\;
}
$\Lambda'\gets\Lambda'\cup\Lambda$\;
$\Lambda\gets\emptyset$\;
$\pi\gets$ element indexes of $S$ in increasing $w_1(\cdot)$ order\;
\For{$i\in\{1,\ldots,|S|-1\}$}{
$\delta_1,\delta_2\gets w_1S_{\pi(i+1)}-w_1S_{\pi(i)},w_2S_{\pi(i)}-w_2S_{\pi(i+1)}$\;
$\Lambda\gets\Lambda\cup\{\delta_1/(\delta_1+\delta_2)\}$\;
}
}
\caption{Special case of Algorithm \ref{alg:convex_fillkd} for $k=2$}
\label{alg:convex_fill2d}
\end{algorithm}

\section{MOEA/D with weight scalarization}
Multi-Objective Evolutionary Algorithm based on Decomposition (MOEA/D), introduced in \cite{QingfuZhang2007}, is a co-evolutionary framework characterized by simultaneous optimization of single-objective subproblems in a multi-objective problem. While there are many approaches to decompose the multi-objective into single-objectives, we consider the classical approach that is weight scalarization \cite{Ishibuchi1998}, as hinted in preceding sections. This simple scheme is sufficient in approximating $F$ and even enumerating $\Conv(F)$.

\subsection{Description}
MOEA/D uses two fitness functions, a scalar function formulated by the decomposition scheme and a vector function for dominance checking \cite{QingfuZhang2007}. To account for the matroid base constraint, we use the penalty term formulated in \cite{Reichel2008}, which was adapted from prior work on MST \cite{Neumann2007}. Letting $w_{max}:=\max_{(i,e)\in\{1,\ldots,k\}\times E}w_{i,e}$, we have the fitness $f_\lambda$ of $x\in\{0,1\}^m$ at trade-off $\lambda$, and the fitness vector $g$ for dominance checking where $\mathbb{1}$ is the one vector.
\begin{align}\label{eq:weighted_fitness}
f_\lambda(x):=m(n-r(x))w_{max}+w^{(\lambda)}x,\quad g(x):=m(n-r(x))w_{max}\mathbb{1}+wx
\end{align}

The MOEA/D for the MOMWB problem is outlined in Algorithm \ref{alg:moead}. The fitness functions defined in Eq. \eqref{eq:weighted_fitness} and the input trade-off set realize the decomposition, and the algorithm evolves a population for each scalarized subproblem with potential neighborhood-based collaboration. During the search, it maintains a non-dominated solution set $S$, which does not influence the search and is returned as output. An optimum to each scalarized subproblem is a supported solution. Note in this formulation, MOEA/D keeps ties in each subproblem, allowing all found optima to participate in mutation operations. This is to avoid having to flip more than two bits to jump from a supported solution to an uncollected point in $\Conv(F)$. We will see that while this may increase the population size, it does not affect the run-time to reach optimality in each subproblem.
\begin{algorithm}[t]
\KwIn{A MOMWB instance, trade-off set $\Lambda$, neighborhood size $N\geq1$}
\KwOut{$S$}
$\forall\lambda\in\Lambda,B_\lambda\gets N$ nearest neighbors of $\lambda$ in $\Lambda$ (Euclidean distance)\;
$\forall\lambda\in\Lambda,P_\lambda\gets$ a random sample from $\{0,1\}^m$\;
$S\gets\emptyset$\;
\While{stopping conditions not met}{
%\itemindent=0pc
\For{$\lambda\in\Lambda$}{
$x\gets$ uniformly sampled from $P_\lambda$\;
$y\gets$ independent bit flips on $x$ with probability $1/m$\;
$D\gets\{l\in B_\lambda:\forall z\in P_l,f_{l}(y)<f_{l}(z)\}$\;
$T\gets\{l\in B_\lambda:\forall z\in P_l,f_{l}(y)=f_{l}(z)\}$\;
$\forall l\in D,P_{l}\gets\{y\}$\;
$\forall l\in T,P_{l}\gets P_{l}\cup\{y\}$ removing solutions with duplicate images\;\label{line:remove_dup}
$S\gets$ non-dominated individuals in $S\cup\{y\}$ under relation $\preceq_g$\;
}
}
\caption{MOEA/D for MOMWB}
\label{alg:moead}
\end{algorithm}

\subsection{Expected time to minimize scalarized weights}\label{sec:runtime_approx}
In the following, we do not assume a particular value of $N$, and the results hold for any $N\geq1$. Furthermore, we exclude the trivial instances with $n=m$ and $n=0$, which admit exactly one base each.
\begin{lemmaE}[][category=runtime,end]\label{lemma:initial_base}
MOEA/D working ontrade-off set $\Lambda$ finds a base's superset for each $\lambda\in\Lambda$ within $O(|\Lambda|m\log n)$ expected search points.
\end{lemmaE}
\begin{proof}
First, we observe that a set supersets a base iff its rank is $n$. We see that for all $\lambda\in[0,1]$ and $x,y\in\{0,1\}^m$, $r(x)>r(y)$ implies $f_\lambda(x)<f_\lambda(y)$. Thus, for each $\lambda\in\Lambda$, MOEA/D performs (1+1)-EA search toward a base's superset with fitness $f_\lambda$, which concludes in $O(m\log n)$ expected steps \cite{Reichel2008}. The claim follows from the fact that MOEA/D produces $|\Lambda|$ search points in each step.
\end{proof}

Let $OPT_\lambda$ be the optimal value to $(E,r,w^{(\lambda)})$, we have the following drift argument proven in \cite{Reichel2008} for the standard bit mutation in the MWB problem.
\begin{lemma}[\cite{Reichel2008}, Proposition 9]\label{lemma:progress}
Given a trade-off $\lambda\in[0,1]$ and $x\in\{0,1\}^m$, if $x$ supersets a base, then there are $n$ 2-bit flips and $m-n$ 1-bit flips on $x$ reducing $f_\lambda(x)$ on average by $(f_\lambda(x)-OPT_\lambda)/m$.
\end{lemma}

We use the same ideas as the proof of Theorem 2 in \cite{Reichel2008}, while sharpening an argument to derive a slightly tighter bound.
\begin{theoremE}[][category=runtime,end]\label{theorem:meta_runtime}
MOEA/D working ontrade-off set $\Lambda$ finds MWBs to instances scalarized by trade-offs in $\Lambda$ in $O\left(|\Lambda|m\log n+\sum_{\lambda\in\Lambda}m^2(\log(m-n)+\log w_{max}+\log d_\lambda)\right)$ expected search points where $d_\lambda:=\min\{a>0:a\lambda,a(1-\lambda)\in\mathbb{N}\}$.
\end{theoremE}
\begin{proof}
We assume each solution in $P_\lambda$ supersets a base for all $\lambda\in\Lambda$; this occurs within expected time $O(|\Lambda|m\log n)$, according to Lemma \ref{lemma:initial_base}. Since for each $\lambda\in\Lambda$, the best improvement in $f_\lambda$ is retained in each step, the expected number of steps MOEA/D needs to minimizes $f_\lambda$ is at most the expected time (1+1)-EA needs to minimizes $f_\lambda$. We thus fix a trade-off $\lambda$ and assume the behaviors of (1+1)-EA. Note that we use $d_\lambda\cdot w^{(\lambda)}$ in the analysis instead for integral weights; we scale $f_\lambda$ and $OPT_\lambda$ accordingly.

We call the bit flips described in Lemma \ref{lemma:progress} \emph{good flips}. Let $s$ be the current search point, if good 1-bit flips incur larger total weight reduction than good 2-bit flips on $s$, we call $s$ 1-step, and 2-step otherwise. If at least half the steps from $s$ to the MWB $z$ are 1-steps, Lemma \ref{lemma:progress} implies the optimality gap of $s$ is multiplied by at most $1-1/2(m-n)$ on average after each good 1-bit flip. Therefore, from $f_\lambda(s)\leq d_\lambda(m-n)w_{max}+OPT_\lambda$, the expected difference $D_L$ after $L$ good 1-bit flips is $E[D_L]\leq d_\lambda(m-n)w_{max}(1-1/2(m-n))^L$. At $L=\lceil(2\ln 2)(m-n)\log(2d_\lambda(m-n)w_{max}+1)\rceil$, $E[D_L]\leq1/2$ and by Markov's inequality and the fact that $D_L\geq0$, $\Pr[D_L<1]\geq1/2$. Since weights are integral, $D_L<1$ implies that $z$ is reached. The probability of making a good 1-bit flip is $\Theta((m-n)/m)$, so the expected number of steps before $L$ good 1-bit flips occur is $O(Lm/(m-n))=O(m(\log(m-n)+\log w_{max}+\log d_\lambda))$. Since 1-steps take up most steps between $s$ and $z$, the bound holds.

If at least half the steps from $s$ to $z$ are 2-steps, Lemma \ref{lemma:progress} implies the optimality gap of $s$ is multiplied by at most $1-1/2n$ on average after each good 2-bit flip. Repeating the argument with $L=\lceil(2\ln 2)n\log(2d_\lambda(m-n)w_{max}+1)\rceil$ and the probability of making a good 2-bit flip being $\Theta(n/m^2)$, we get the bound $O(m^2(\log(m-n)+\log w_{max}+\log d_\lambda))$. Summing this over all $\lambda\in\Lambda$ gives the total bound.
\end{proof}

In order for MOEA/D to reach $k$-approximation and not lose it afterward, it suffices that $\Lambda$ is sufficient and each scalarized subproblem admits optima with a unique image. As mentioned and from Lemma \ref{lemma:tie_partition}, this can be obtained by sampling from the interiors of convex polytopes in $H_A$. For $k=2$, this can be done by taking a complete scalar trade-off set $A$ (e.g. as returned by Algorithm \ref{alg:convex_fill2d}) and include $(a+b)/2$ (which is an interior point) for each non-empty interval $(a,b)$ bounded by consecutive elements in $A\cup\{0,1\}$. Under the integer weights assumption, this method gives rational trade-offs whose integral denominators are $O(w_{max}^2)$, so we have the following bound from Corollary \ref{corollary:2obj_extreme_bound}.

\begin{corollary}
For a bi-objective instance, MOEA/D working on a minimal sufficient trade-off set finds a sufficient solution set within $O(m^2\sqrt{\min\{n,m-n\}}(m(\log(m-n)+3\log w_{max})+\log n))$ expected number of search points.
\end{corollary}
 
This method can be generalized to higher dimensions. Instead of taking an average of two consecutive elements, we can take the average of the vectors normal to $k$ facets (i.e. $(k-1)$-dimensional faces) of $\Conv(F)$ that meet at an extreme point. Since each facet is determinable by $k$ points with integral coordinates, each such ($L_1$-normalized) vector admits rational coordinates with denominator at most $kw_{max}$. Therefore, the trade-offs derived by this method admit rational representations whose denominators are $O(k^{k+1}w_{max}^k)$, giving the run-time upper bound from Corollary \ref{collorary:sufficient_no_upperbound} under the assumption that $k$ is sufficiently small.
\begin{corollary}
Given a $k$-obbjective instance where $k\in o(m)$ and $k\in o(w_{max})$, MOEA/D working on a minimal sufficient trade-off set guarantees $k$-approximation within $O(m^{2k-1}(m(\log(m-n)+(k+1)\log w_{max})+\log n))$ expected number of search points.
\end{corollary}
As a side note, since MOEA/D uses standard bit mutation, we can replace $w_{max}$ with $m^m$ and remove $\log d_\lambda$ from the bound in Theorem \ref{theorem:meta_runtime} to arrive at weight-free asymptotic bounds \cite{Reichel2009}.

\subsection{Expected time to enumerate Conv($F$)}
We see from Corollary \ref{corollary:2bit_enum} that MOEA/D with a complete trade-off set can collect all points in $\Conv(F)$ with 2-bit flips starting from an optimum to each subproblem. As mentioned, this is afforded by allowing all found optima to undergo mutation.
\begin{theoremE}[][category=runtime,end]
Assuming distinct supported solutions have distinct images under $w$, MOEA/D working on a minimal complete trade-off set $\Lambda$, and starting from an optimum for each $\lambda\in\Lambda$ enumerates $C:=\Conv(F)$ in $O(|\Lambda||C|^2m^{2})$ expected number of search points.
\end{theoremE}
\begin{proof}
From Corollary \ref{corollary:2bit_enum}, to collect a new point in $C$, it is sufficient to perform a 2-bit flip on some supported solution. In worst-case, there can be only one trade-off $\lambda\in\Lambda$ such that all non-extreme supported solutions minimize $w^{(\lambda)}$, so the correct solution is mutated with probability at least $1/l$ in each iteration, where $l$ is the number of already collected points. Since $|\Lambda|$ search points are generated in each iteration, the expected number of search points required to enumerate $C$ is $O(|\Lambda|m^2\sum_{l=1}^{|C|}l)=O(|\Lambda||C|^2m^{2})$.
\end{proof}
With this, Theorem \ref{theorem:meta_runtime} and Corollary \ref{corollary:2obj_extreme_bound} and \ref{collorary:sufficient_no_upperbound}, we have the following expected run-time bounds under the distinct image assumption. Note this assumption can be removed by having MOEA/D keep duplicate images at line \ref{line:remove_dup}.
\begin{corollary}
For a bi-objective instance, MOEA/D working on a minimal complete trade-off set enumerates $C:=\Conv(F)$ in expected time $O(m^{2}\sqrt{\min\{n,m-n\}}(m(\log(m-n)+3\log w_{max}+|C|^2)+\log n))$.
\end{corollary}
\begin{corollary}
Given a $k$-objective instance where $k\in o(m)$ and $k\in o(w_{max})$, MOEA/D working on a minimal complete trade-off set enumerates $C:=\Conv(F)$ in expected time $O(m^{2k-1}(m(\log(m-n)+(k+1)\log w_{max}+|C|^2)+\log n))$.
\end{corollary}

\section{Experimental investigation}
In this section, we perform computational runs of MOEA/D on various bi-objective minimum spanning tree instances. Spanning trees in a connected graph $G=(V,E)$ are bases of the graphic matroid defined by $G$ admitting the ground set $E$. The rank of such a matroid (i.e. the size of the spanning tree) equals $|V|-1$ and its rank function is defined with $r(x)=|V|-cc(x)$ where $cc(x)$ is the number of connected components in $(V,x)$. In notations, we have $n=|V|-1$ and $m=|E|$. We use simple undirected graphs in our experiments, and the edge-set representation of solutions in the implementations of the algorithms \cite{Raidl}.

\subsection{Setting and performance metrics}
We uniformly sample graphs with $|V|\in\{26,51,101\}$ and $|E|\in\{150,300\}$. In this procedure, edges are added randomly into an empty graph up to the desired edge count, and this is repeated until a connected graph is obtained. Each edge weight is an integer sampled independently from $\mathcal{U}(1,100)$. We generate two weighted graphs with each setting, making 12 instances in total.

With this experiment, we aim to measure the algorithms' performances in finding solutions mapped to all extreme points, we denote this set of target points with $R$. We compare MOEA/D against GSEMO, previously studied for its performance in bi-objective MST \cite{Neumann20071}. For GSEMO, we use the fitness function $g$ defined in Eq. \eqref{eq:weighted_fitness}. Since GSEMO checks for dominance in each iteration across its entire population, we set $N:=|\Lambda|$ for MOEA/D to match. Here, the input $\Lambda$ to MOEA/D is derived from a complete trade-off set output by Algorithm \ref{alg:convex_fill2d} in the manner described in Section \ref{sec:runtime_approx}. The points given by Algorithm \ref{alg:convex_fillkd} are the target points $R$, and each run is terminated if all target points are hit. Additionally, each run is terminated after at most $\lceil 3|R|m^2\log(m-n)\rceil$ evaluations (each evaluation is a call to function $g$). Each algorithm is run on each instance 10 times. Their performances are measured with the followings ($S$ is the final population returned by the algorithm):
\begin{itemize}
\item\textbf{Success rate:} The number of runs where all target points are hit within the evaluation budget.
\item\textbf{Cover rate:} The proportion of hit target points after termination, $|R\cap\{wx:x\in S\}|/|R|$. A run is successful if this reaches 100\%.
\item\textbf{Modified inverted generational distance (IGD+):} The distance between the output and the target points \cite{Ishibuchi2015}, $\sum_{y\in R}\min_{x\in S}\sqrt{\sum_{i=1}^k(\max\{(wx)_i-y_i,0\})^2}/|R|$. A run is successful if this reaches 0.
\item\textbf{T:} The number of evaluations until all target points are hit.
\end{itemize}
We remark that the fitnesses $f_\lambda$ can be quickly computed alongside $g$, incurring minimal overheads. In fact, the run-time bottleneck is in checking the number of connected components.

\subsection{Experimental results}
The results are shown in Table \ref{tab:results_main}, with IGD+ and cover rate from MOEA/D omitted due to them being 0 and 100\% across all instances, respectively. These are contextualized by the listed instance-specific parameters. Of note is the number of target points $|R|$ which is smaller than the upper bound in Theorem \ref{theorem:2obj_extreme_bound} in all instances.

Immediately, we see the GSEMO failed to hit all target points within the evaluation budget in most runs, while MOEA/D succeeded in every run. In most cases, GSEMO hit at most one target point. Inspecting the output points and IGD+ reveals that its population converged well to the non-dominated front, yet somehow misses most extreme points. In contrast, MOEA/D hit all target points within up to 92\% of the evaluation budget, though there are significant relative variances in the run-times.

Inspecting the run-times of MOEA/D in relation to the evaluation budgets, we see that the means of ratios remain fairly stable across instances. This suggests the asymptotic bound in Theorem \ref{theorem:meta_runtime} is not overly pessimistic. Instances 5 and 6 are particularly interesting as they are ostensibly the easiest due to the small number of extreme points, yet MOEA/D seems to require the most fractions of the budget. Given that these instances exhibit the smallest $m-n$, this can be explained by the interference of lower-order terms in the asymptotic bound, which are not counted in the budgets.

\begin{table*}[t!]
\centering
\caption{Means and standard deviations of performance statistics from GSEMO and MOEA/D on bi-objective MST instances. Means and standard deviations of $T$ are computed over successful runs only. All differences are statistically significant.}\vspace{-5pt}
\label{tab:results_main}
% \small
% \renewcommand{\arraystretch}{.8}
\begin{tabular}{lllll|@{\hspace{0.4em}}c@{\hspace{0.5em}}c@{\hspace{0.6em}}c@{\hspace{0.6em}}c@{\hspace{0.4em}}|@{\hspace{0.4em}}c@{\hspace{0.5em}}c}\toprule
\multirow{2}{*}{Id}&\multirow{2}{*}{$m$}&\multirow{2}{*}{$n$}&\multirow{2}{*}{$|R|$}&\multirow{2}{*}{max eval.}&\multicolumn{4}{@{\hspace{0.4em}}c@{\hspace{0.4em}}|@{\hspace{0.4em}}}{GSEMO}&\multicolumn{2}{c}{MOEA/D}\\\cmidrule(lr){6-11}&&&&&Suc. rate&Cover rate&IGD+&T/max eval.&Suc. rate&T/max eval.\\\midrule1&150&25&39&12710536&2/10&2.5$\pm$0.06\%&0.13$\pm$0.11&79$\pm$11\%&10/10&50$\pm$15\%\\2&150&25&31&10103247&0/10&3.1$\pm$0.14\%&0.25$\pm$0.32&N/A&10/10&43$\pm$11\%\\3&150&50&45&13988205&0/10&1.9$\pm$0.1\%&1.7$\pm$0.7&N/A&10/10&52$\pm$11\%\\4&150&50&45&13988205&0/10&2$\pm$0.083\%&0.92$\pm$0.23&N/A&10/10&47$\pm$12\%\\5&150&100&35&9242155&0/10&2.5$\pm$0.13\%&2.6$\pm$1.2&N/A&10/10&63$\pm$15\%\\6&150&100&36&9506216&0/10&2.2$\pm$0.21\%&3.7$\pm$2.3&N/A&10/10&58$\pm$11\%\\7&300&25&45&68243769&3/10&2.2$\pm$0.031\%&0.06$\pm$0.072&86$\pm$13\%&10/10&41$\pm$11\%\\8&300&25&49&74309882&1/10&2$\pm$0.031\%&0.076$\pm$0.083&94$\pm$0\%&10/10&40$\pm$8.1\%\\9&300&50&66&98392434&0/10&1.4$\pm$0.048\%&0.49$\pm$0.13&N/A&10/10&50$\pm$11\%\\10&300&50&63&93920051&0/10&1.5$\pm$0.045\%&0.8$\pm$0.24&N/A&10/10&60$\pm$15\%\\11&300&100&79&113013110&0/10&1.1$\pm$0.043\%&2$\pm$0.53&N/A&10/10&60$\pm$8.3\%\\12&300&100&80&114443656&0/10&1.1$\pm$0.072\%&3.1$\pm$1&N/A&10/10&58$\pm$11\%\\\bottomrule
\end{tabular}
\end{table*}
\section{Conclusion}
In this study, we contribute to the theoretical analyses of evolutionary multi-objective optimization in the context of non-trivial combinatorial problems. We give the first run-time analysis of the MOEA/D algorithm for a broad problem class that is multi-objective minimum weight base problem. In particular, we show a fixed-parameter polynomial expected run-time for approximating the non-dominated front, simultaneously extending existing pseudo-polynomial bounds for GSEMO to arbitrary number of objectives and broader combinatorial structures. Our experiments in random bi-objective minimum spanning tree instances indicate that MOEA/D significantly outperforms GSEMO in the computing extreme points under an appropriate decomposition. Along the way, we prove properties that give further insight into the problem of interest.

\section*{Acknowledgements}
%This work was supported by the %Australian Research Council by %grants
%DP190103894 and FT200100536. %A.\ M.\ Sutton was supported by %NSF grant 2144080.

This work was supported by Australian Research Council grants
DP190103894 and FT200100536, and by National Science Foundation grant 2144080.

\bibliographystyle{abbrv}
\bibliography{refs}
\end{document}